\documentclass[11pt,a4paper]{article}
\usepackage[hyperref]{ranlp2023}
\usepackage{times}
\usepackage{latexsym}

\usepackage{microtype}
\usepackage{graphicx, subfigure}
\usepackage{algorithm}
\usepackage{algpseudocode}
\usepackage{makecell, multirow, multicol, amsmath}
\usepackage{hyperref,url}
\usepackage{tcolorbox}

\aclfinalcopy 

\title{The Erosion of LLM Signatures: Can We Still Distinguish Human and LLM-Generated Scientific Ideas After Iterative Paraphrasing?}

\author{
Sadat Shahriar, Navid Ayoobi, Arjun Mukherjee \\
University of Houston, Texas, USA \\
\texttt{sadat.shrr@gmail.com, nayoobi@cougarnet.uh.edu, arjun@cs.uh.edu}
}

\date{}

\begin{document}
\maketitle
\begin{abstract}
With the increasing reliance on LLMs as research agents, distinguishing between LLM and human-generated ideas has become crucial for understanding the cognitive nuances of LLMs' research capabilities. While detecting LLM-generated text has been extensively studied, distinguishing human vs LLM-generated \textit{scientific ideas} remains an unexplored area. In this work, we systematically evaluate the ability of state-of-the-art (SOTA) machine learning models to differentiate between human and LLM-generated ideas, particularly after successive paraphrasing stages. Our findings highlight the challenges SOTA models face in source attribution, with detection performance declining by an average of 25.4\% after five consecutive paraphrasing stages. Additionally, we demonstrate that incorporating the research problem as contextual information improves detection performance by up to 2.97\%. Notably, our analysis reveals that detection algorithms struggle significantly when ideas are paraphrased into a simplified, non-expert style, contributing the most to the erosion of distinguishable LLM signatures. 
\end{abstract}

\section{Introduction}

\begin{figure*}[h]
\centering
  \includegraphics[width=.8\textwidth]{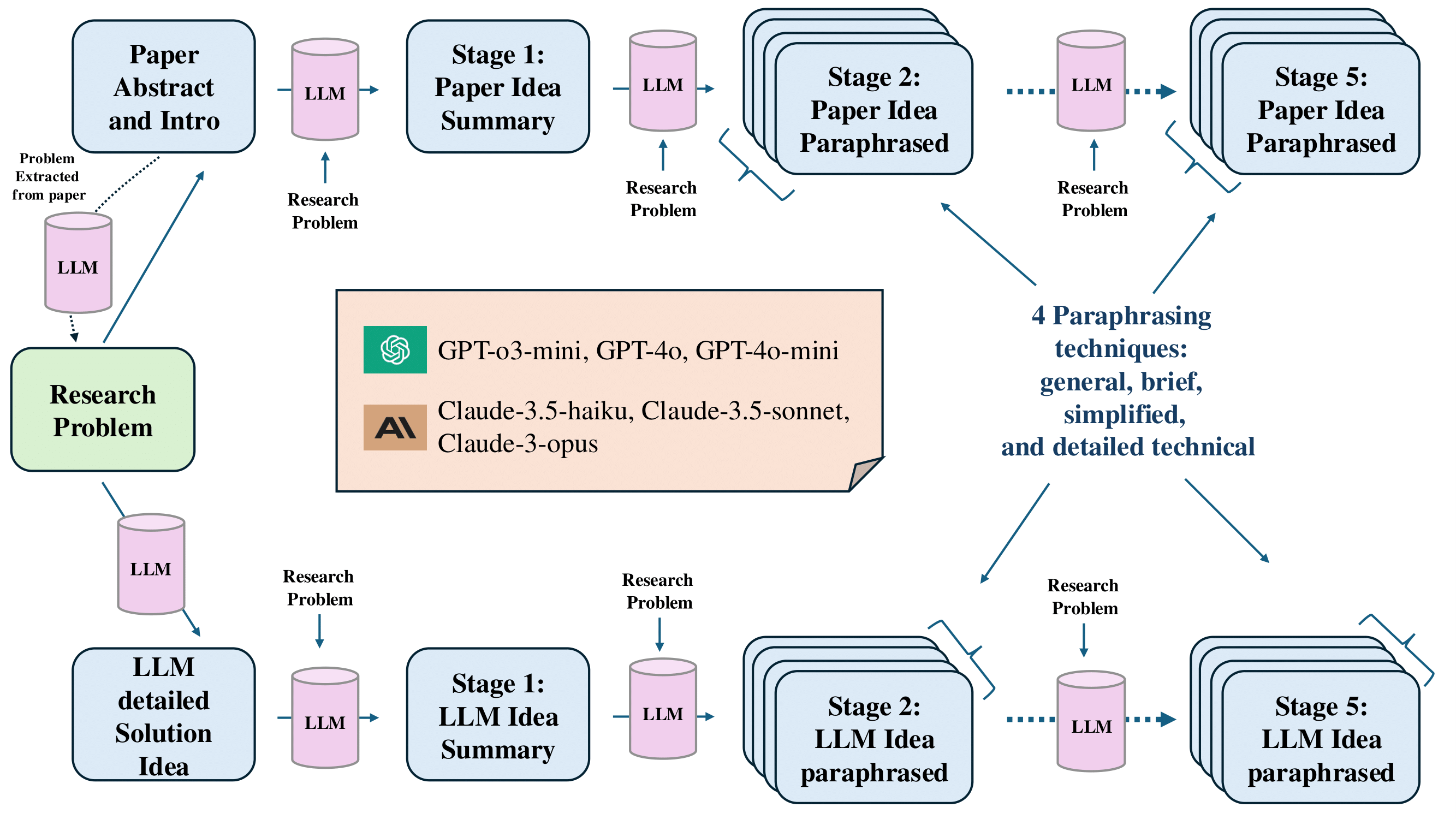}
  \caption{Idea Generation and Paraphrasing Workflow: The process begins with extracting the Research Problem from papers and then generate corresponding scientific ideas using six different LLMs. Both human and LLM-generated ideas are first summarized and subsequently paraphrased across five stages using four distinct paraphrasing techniques (To reduce visual clutter and redundancy, we abstracted Stages 3 and 4, as they represent similar paraphrasing strategies). }
  \label{fig:experiments}
\end{figure*}

Recent advances in LLMs have demonstrated extraordinary capabilities extending far beyond mundane conversational tasks \cite{boiko2023emergent, zhao2023survey}. Notably, these models can even engage in complex cognitive activities traditionally reserved for human intellect, such as hypothesis generation, reasoning, and scientific inquiry \cite{boiko2023emergent, si2024can}. This remarkable development raises a fundamental question: Given humanity's millennia-long tradition of knowledge creation and  dissemination-- and the subsequent encoding into vast linguistic datasets: can we still reliably discern whether novel ideas originate from humans or are algorithmically produced by LLMs?

Si et al. showed that LLMs can generate more novel ideas compared to human experts, though these ideas are not always practically feasible \cite{si2024can}. While novelty definitions carry inherent subjectivity, on a broader scale, LLMs still exhibit significant capability in producing innovative research ideas. As such, distinguishing between ideas generated by LLMs vs humans becomes increasingly important, as it provides deeper insights into LLM cognitive patterns, ensures academic integrity, and aids in maintaining transparency by clearly attributing authorship, ultimately influencing trust in scholarly contributions and guiding responsible AI deployment in research contexts.


While prior research on detecting LLM-generated text has focused on watermarking \cite{zhao2023protecting}, zero-shot methods \cite{yang2023dna, mitchell2023detectgpt}, and fine-tuned classifiers \cite{hu2023radar}, our study takes a fundamentally different approach. Rather than identifying LLM-generated \textit{text}, we examine the resilience of \textit{ideas}--which persist beyond surface-level writing styles. Unlike text, ideas are conceptually immutable; a human-conceived idea remains human in essence, even if heavily paraphrased by an LLM. We investigate whether these underlying origins: human or LLM—remain detectable after successive paraphrasing and stylistic transformations. To the best of our knowledge, this is the first study to explore scientific idea attribution in such a nuanced and dynamic setting.

Ideas manifest across diverse contexts, but in this research, we define an ``idea'' specifically as a proposed solution addressing a given research problem, using `scientific idea' and `idea' interchangeably. Scientific ideas inherently reflect nuanced thinking and careful planning, which distinguishes them from mere linguistic outputs. Formally, given a research problem \(RP\), an idea can be represented as a response \( r = f(RP) \), where \(f\) denotes either human or LLM generation. To evaluate whether the essence of human or LLM-generated ideas persists through stylistic variations, we iteratively paraphrase these ideas through multiple stages. At each paraphrasing stage \( n \), the idea transforms as \( r_n = f_{pn}(r_{n-1}, RP) \). Paraphrasing serves two critical purposes: firstly, in real-world scenarios, ideas are communicated through varied expressions and settings—yet retain their core meaning; secondly, without paraphrasing, classifiers might easily identify the source due to stylistic cues specific to scientific paper writing, conflating stylistic detection with genuine idea detection.


In this research, We collect 846 papers from five top CS conferences to extract their main research problems. We then prompt LLMs to generate original ideas for each problem. Human-generated (from papers) and LLM-generated ideas undergo systematic summarization and multi-stage paraphrasing using four strategies: general paraphrase, simplified summary, brief summary, and detailed technical paraphrase. Figure \ref{fig:experiments} illustrates this workflow.

We employ SOTA classifiers to assess detection performance across paraphrasing stages, revealing an average decline of 25.4\% from Stage 1 to Stage 5. This deterioration suggests that characteristic ``LLM signatures'' initially present in earlier stages--such as specific word choices, linguistic patterns, or stylistic markers—gradually diminish through successive paraphrasing. As these superficial markers fade, traditional text-based classifiers increasingly struggle to differentiate between human and LLM-generated ideas.

Our main contributions are as follows:
\begin{itemize}
\item We create and release a comprehensive dataset consisting of original and multi-stage paraphrased scientific ideas, systematically generated using cutting-edge LLMs.

\item Through extensive evaluation using various classification algorithms, we empirically demonstrate the inherent challenges involved in identifying LLM-generated ideas, particularly as these ideas undergo iterative paraphrasing and stylistic transformations.

\end{itemize}

\section{Related Works}

LLMLu et al. introduced \textit{AI Scientist}, an end-to-end framework designed for scientific discovery using LLMs. This framework autonomously generates novel research ideas, implements experimental code, executes experiments, visualizes results, composes scientific papers, and even simulates a peer-review process to evaluate its findings \cite{lu2024ai}. Similarly, Baek et al. proposed a research agent capable of automatically formulating problems, suggesting methods, and designing experiments. Their approach iteratively refines these elements through feedback provided by collaborative LLM-powered reviewing agents \cite{baek2024researchagent}.

Li et al. developed specialized LLM-driven agents, namely \textit{IdeaAgent} and \textit{ExperimentAgent}, tailored for research idea generation, experimental implementation, and execution within the machine learning domain \cite{li2024mlr}. OpenAI’s Deep Research initiative demonstrated the potential of LLMs to gather, analyze, and synthesize extensive online resources, producing comprehensive reports like those of human research analysts \cite{openai2025deepresearch}.

Si et al. conducted a comparative study on the novelty of ideas generated by expert NLP researchers and LLM ideation agents \cite{si2024can}. Their results suggested that LLMs can generate ideas that surpass human-generated ideas in terms of novelty. However, their approach involved prompting-based idea generation across predefined NLP topics. In contrast, our research explicitly defines an idea as a response to a given research problem. This definition allows for a more unbiased comparison between human and LLM-generated ideas and reduces topic-related bias.

Unlike previous studies primarily focused on enabling research potential of LLMs, and comparing novelty, our work shifts attention toward the challenge of detection. Specifically, we investigate the inherent difficulty of distinguishing human-generated ideas from those produced by LLMs, especially as these ideas undergo multiple stages of paraphrasing.

\section{Methodology}
To generate research ideas, we first extract research problems from scientific papers and feed these into LLMs. Subsequently, we apply cascading paraphrasing to both human-written and LLM-generated ideas. Finally, we evaluate the distinguishability of these ideas at each paraphrasing stage using several SOTA classifiers.
\subsection{Data Collection}
\begin{table}[ht]
\centering
\begin{tabular}{lccccc}
\hline
        & 2017 & 2018 & 2019 & 2020 & 2021 \\
\hline
ACL              & 19   & 15   & 12   & 35   & 19   \\
EMNLP            & 13   & 27   & 27   & 43   & 38   \\
ICLR             & 1    & 15   & 15   & 29   & 39   \\
ICML             & 28   & 24   & 34   & 32   & 51   \\
NeurIPS          & 37   & 39   & 56   & 101  & 97   \\
\hline
\end{tabular}
\caption{Conference counts by publication year.}
\label{tab:conference_counts}
\end{table}

To compile our dataset, we first sample 846 scientific papers from a larger collection drawn from five A*-rated computer science conferences---ACL, EMNLP, ICLR, ICML, and NeurIPS---spanning 2017 to 2021~\cite{core_portal}. This sample size was chosen primarily due to the substantial computational and financial resources required for large-scale generation and extensive cascading paraphrasing using SOTA LLM APIs. Table~\ref{tab:conference_counts} summarizes the sampled dataset, while detailed statistics are available in our repository.\footnote{Check the Appendix of the full paper: \url{https://github.com/sadat1971/Erosion_LLM_Signatures/blob/main/Paper/RANLP__LLMErosion_cameraReady.pdf}} We include only papers published up to 2021 to ensure the integrity of our analysis, as this guarantees that the ideas originate purely from humans, predating the release of ChatGPT in 2022~\cite{OpenAI2025ChatGPT}.

\subsection{Extracting Research Problem}

We extract the research problem from the first two pages of each paper, selecting five different LLMs at random for each extraction. In general, these pages encompass the abstract and introduction, where problem statements are typically presented either explicitly or implicitly. To minimize the risk of LLMs incorporating elements of the solution, we explicitly prompt them to focus solely on the problem itself (find the prompts in Appendix) \footnotemark[\value{footnote}].

\subsection{LLM Idea Generation}
The extracted research problem is used as input to the LLM along with carefully designed instructions to generate potential research ideas. This process constitutes the core of LLM-driven idea generation. We employ two distinct prompting strategies. The first is a general prompting approach, where the LLM is simply instructed to provide a detailed research solution. The second approach, inspired by the idea generation technique outlined in \cite{si2024can}, involves a more structured prompt with step-by-step guidance on explaining the methodology, techniques employed, novelty, and contributions. While both approaches yielded comparable results, the latter tends to produce slightly more detailed and descriptive responses. To incorporate both prompting styles, we apply the general prompting method to half of the samples and the structured approach to the remaining half. A detailed description of both prompting strategies is provided in Appendix \footnotemark[\value{footnote}].

\subsection{Idea Paraphrasing} 
Since we want to differentiate ideas generated by humans from those produced by LLMs, a direct comparison between LLM-generated ideas and research papers (first two pages) is not feasible. This is primarily due to the presence of stylistic cues that algorithms can easily detect, as well as inconsistencies in formatting across these two categories. Consequently, distinguishing ideas at this stage would not be reliable.

Hence, we employ a multi-stage cascade of summarization and paraphrasing. In the first stage (Stage 1), we generate a three-paragraph summary of both the first two pages of each paper and the corresponding LLM-generated ideas. In the second stage, we apply four distinct paraphrasing strategies to each summary: (i)\textbf{ general paraphrasing}, (ii) \textbf{paraphrasing for a simplified non-expert audience}, (iii) \textbf{brief summarization}, and (iv) \textbf{detailed technical paraphrasing}. This paraphrasing process continues in a cascaded manner across a total of five stages. To prevent excessive compression or the introduction of additional information, we avoid consecutive applications of the same paraphrasing type. Appendix \ref{Appsub:paraphrase} shows the instruction prompts to generate these paraphrases. 

Through this approach, we obtain 846 paraphrases in Stage 1. From Stages 2 to 5, this expands to 3,384 paraphrases for both LLM-generated and human-written ideas. In total, our process yields 28,764 paraphrased versions of research ideas. One of the authors also manually verified a 1\% of the samples across all paraphrasing stages to ensure consistency. 

\subsection{Generative LLMs}
We utilize six best-performing LLMs to generate data for research problem extraction, idea generation, and the five stages of idea paraphrasing. To ensure optimal performance, we conduct small-scale experiments and manual evaluations of the quality of generated outputs across different LLMs. Additionally, we consider the cost of API usage as a factor in model selection. Based on these trade-offs, we selected three models from OpenAI \cite{openai2025models} and three from Anthropic \cite{Anthropic2025Claude}.

From OpenAI’s suite of models, we use \textbf{GPT-4o}, \textbf{GPT-4o-mini}, and \textbf{O3-mini}. From Anthropic, we employ \textbf{Claude-3.5-Haiku}, \textbf{Claude-3.5-Sonnet}, and \textbf{Claude-3-Opus}. Across all stages of our study, 63\% of the data was generated using OpenAI’s models, while the remaining 37\% was produced using Anthropic’s models. Table \ref{tab:LLM-proportion} presents the exact distribution of data generation across the selected LLMs.

To minimize topic bias, we ensured that the same LLM was used for both summarizing the research paper and generating the summary of the corresponding LLM-generated idea. This consistency was maintained throughout the paraphrasing process as well. 

\begin{table}[h]

    \centering
    \begin{tabular}{c|c}
         \textbf{LLM} & \textbf{\% of Data} \\
         \hline
         gpt-4o-mini & 41.37\\
         gpt-4o & 5.91\\
         gpt-o3-mini & 17.73 \\
         claude-3-5-haiku & 23.64\\
         claude-3-5-sonnet & 7.09\\
         claude-3-opus & 4.26\\
         \hline
    \end{tabular}
    \caption{Percentage of data generated by each flagship LLMs}
    \label{tab:LLM-proportion}
\end{table}

\subsection{Classifiers}

We evaluated four fine-tuned language models and four text embedding methods, each coupled with downstream classification layers. First, we employed BERT (\textit{bert-base-uncased}) as our baseline, owing to its proven ability in capturing bidirectional contextual information \cite{devlin2019bert}. RoBERTa (\textit{roberta-base}), known for its more extensive pretraining, is included as a strong comparative choice \cite{liu2019roberta}. Additionally, BigBird (\textit{BigBird-RoBERTa-base}) is selected due to its efficient handling of long sequences by employing a sparse attention mechanism, thus avoiding the quadratic complexity in traditional transformers \cite{zaheer2020big}. Finally, we incorporate T5 (\textit{t5-base}), a text-to-text transformer featuring an encoder-decoder architecture that fundamentally differs from BERT-style models by translating input text into target text \cite{raffel2020exploring}.

For embedding-based representations, we use the sentence-transformers' \textit{all-MiniLM-L6-v2} as our baseline, encoding text into 384-dimensional vectors \cite{wang2020minilm}. Additionally, we selected three advanced embedding models—\textit{GIST-Embedding-v0} \cite{solatorio2024gistembed}, \textit{gte-base-en-v1.5} \cite{zhang2024mgte}, and \textit{stella\_en\_400M\_v5} \cite{zhang2025jasperstelladistillationsota}, which consist of 109M, 137M, and 435M parameters respectively.

These specific embedding models were chosen based on initial exploratory experiments and by carefully considering the trade-offs between model size and ranking performance \cite{huggingface_mteb2025}. Each embedding representation was subsequently coupled with a downstream two-layer Feed-Forward Neural Network (FFNN).

\section{Experimental Setup}

\begin{table*}[h]
    \centering
    \begin{tabular}{l|ccccc|cccccc}
    \hline
& \multicolumn{5}{c|}{Single Stage Training} & \multicolumn{5}{c}{Combined Stage Training} \\
\hline
& S1 & S2 & S3 & S4 & S5 & S1 & S2 & S3 & S4 & S5 \\
\hline
BERT & 85.8 & 74.1 & 67.1 & 63.2 & 61.1 & 80.7 & 72.6 & 66.1 & 62.4 & 61.2 \\
RoBERTa & 88.1 & 74.4 & 70.3 & 63.8 & 62.7 & 84.1 & 75.2 & 70.8 & 66.7 & 63.9 \\
T5 & 84.2 & 65.4 & 59.0 & 54.4 & 49.9 & 89.7 & 81.0 & 72.1 & 67.1 & 64.0 \\
Bigbird (RP+idea) & \textbf{92.3} & \textbf{83.4} & \textbf{70.9} & \textbf{65.1} & \textbf{63.2} & \textbf{90.5} & \textbf{81.4} & \textbf{72.2} & \textbf{67.2} & \textbf{64.9} \\
MiniLM +FFNN (idea) & 81.2 & 66.6 & 59.3 & 57.0 & 55.1 & 75.2 & 64.2 & 59.6 & 57.2 & 56.0 \\
MiniLM +FFNN (RP+idea) & 83.2 & 69.4 & 61.6 & 58.0 & 56.3 & 75.9 & 66.3 & 60.1 & 60.0 & 57.0 \\
GIST +FFNN (idea) & 83.9 & 71.9 & 64.7 & 60.7 & 53.3 & 78.2 & 70.0 & 64.1 & 61.3 & 58.7 \\
GIST +FFNN (RP+idea) & 86.9 & 73.0 & 65.3 & 61.1 & 59.1 & 78.4 & 70.9 & 64.2 & 61.4 & 59.9 \\
Gte +FFNN (idea) & 89.3 & 72.2 & 63.8 & 61.1 & 56.9 & 78.4 & 66.6 & 62.9 & 59.3 & 56.8 \\
Gte +FFNN (RP+idea) & 89.4 & 73.2 & 64.2 & 61.2 & 57.7 & 79.0 & 69.4 & 63.1 & 59.4 & 57.9 \\
Stella +FFNN (idea) & 90.0 & 73.9 & 65.3 & 61.3 & 56.7 & 77.8 & 70.5 & 64.5 & 61.1 & 58.5 \\
Stella +FFNN (RP+idea) & 91.8 & 77.6 & 67.4 & 63.6 & 58.5 & 85.4 & 74.0 & 66.3 & 64.0 & 60.7 \\
\hline

    \end{tabular}
    \caption{F1-score comparison of various classifiers across different stages (S1 to S5). \textit{Single Stage Training} involves training within a single paraphrase stage, while \textit{Combined Stage Training} aggregates data from all stages for classification.}
    \label{tab:result-table}
\end{table*}

For the experiments, we prepare dataset using a systematic train-test splitting approach to ensure unbiased evaluation. Initially, we have 1,692 samples in Stage 1, comprising equal portions of LLM-generated and human-generated ideas. For subsequent stages (Stage 2 to Stage 5), the dataset expanded to include 6,768 samples, incorporating four distinct paraphrasing styles for each original idea. To avoid data leakage, we perform splits such that there was no overlap between the original solution (and the research problem) in the training and test sets, meaning each problem-solution was exclusive to either the training or testing partition across all stages. This strategy ensures that all paraphrases derived from the same initial research problem statement remain consistently within the same partition, thus maintaining dataset integrity.

We conduct three random train-test splits and report the averaged results across these splits. From each training split, we further allocate 20\% of the data as a validation set, specifically used for hyperparameter tuning. We perform tuning for batch size, number of epochs, dropout rate, and early stopping criteria. 

For all our experiments, we use NVIDIA TITAN RTX (24 GB), Quadro RTX 8000 (48 GB), and NVIDIA GeForce RTX 2080 Ti (11 GB) GPUs. We report the macro F1-score to report the performances.

\section{Results and Discussion}

\begin{figure}[htbp]
    \centering
 \fbox{
    \begin{minipage}{0.44\textwidth}
        \noindent
        \colorbox{green!10!white}{the main idea} is to \colorbox{red!5!white}{develop} \colorbox{red!10!white}{adaptive} \colorbox{green!5!white}{momentum} - \colorbox{red!5!white}{regularized} \colorbox{green!10!white}{federated} \colorbox{red!5!white}{optimization} \colorbox{green!10!white}{(AMRFO)} , a \colorbox{red!5!white}{framework} designed to \colorbox{red!5!white}{enhance} convergence rates and \colorbox{red!5!white}{reduce} \colorbox{green!10!white}{communication} \colorbox{green!10!white}{overhead} in distributed \colorbox{green!10!white}{machine} learning . \colorbox{green!10!white}{AMRFO} achieve this by employing a \colorbox{red!10!white}{multi-stage} \colorbox{red!10!white}{adaptive} \colorbox{red!10!white}{regularization} \colorbox{red!10!white}{mechanism} that includes \colorbox{red!20!white}{adaptive} \colorbox{green!10!white}{momentum} scaling, \colorbox{green!10!white}{stochastic} \colorbox{green!10!white}{communication} \colorbox{red!10!white}{compression}, and \colorbox{red!10!white}{stability}-aware \colorbox{red!10!white}{gradient} \colorbox{red!10!white}{normalization} \colorbox{green!10!white}{thereby} \colorbox{green!20!white}{balancing} communication efficiency , \colorbox{green!10!white}{convergence} \colorbox{green!30!white}{speed}, \colorbox{red!5!white}{and} algorithmic \colorbox{red!10!white}{stability} in \colorbox{green!10!white}{federated} learning \colorbox{red!10!white}{environments}.
    \end{minipage}}
    
    \caption{Integrated Gradients Visualization: Green highlights words that contribute to classifying the text as human-written, while red highlights words that push the classification toward LLM-generated content. The overall text is LLM-idea-summarized}
    \label{fig:Integ-grad}
\end{figure}

\begin{figure}[h]
    \centering
    {\includegraphics[width=0.48\linewidth]{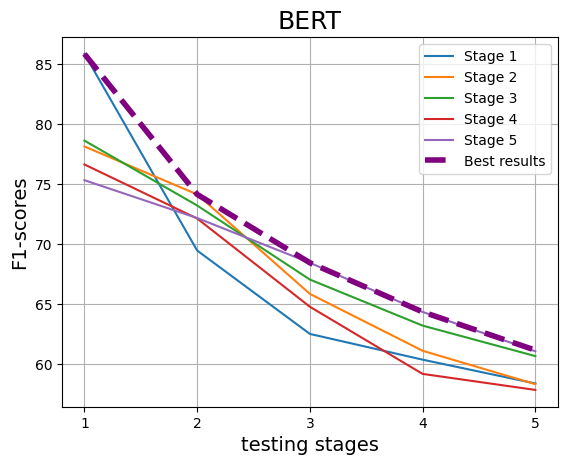}}
    \hfill
 {\includegraphics[width=0.48\linewidth]{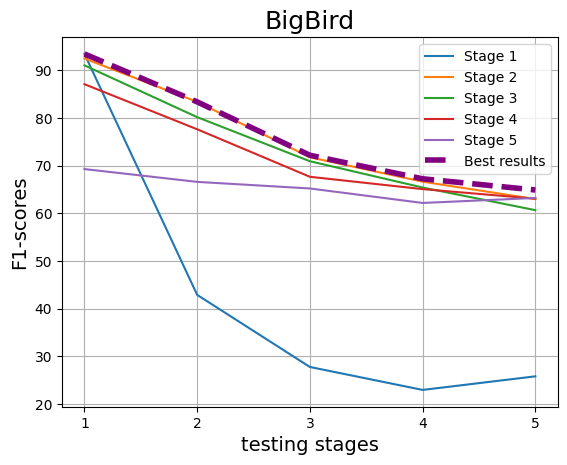}}
    
    \vspace{0.05cm} 
    
    {\includegraphics[width=0.48\linewidth]{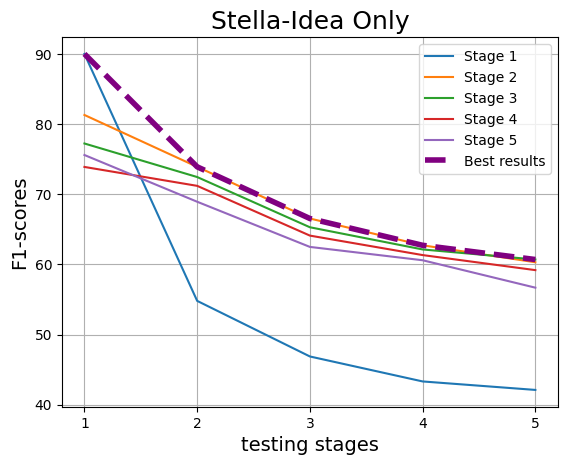}}
    \hfill
    {\includegraphics[width=0.48\linewidth]{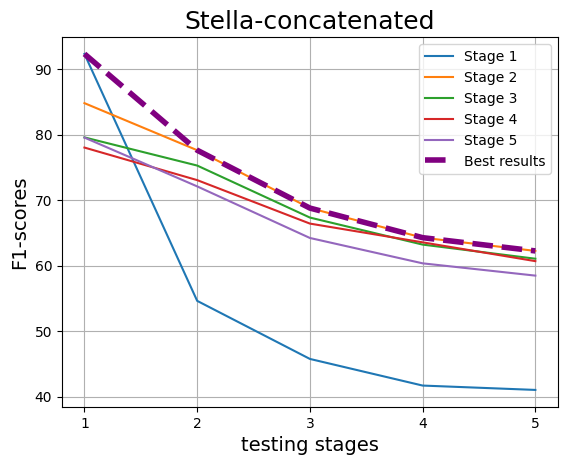}}

    \caption{F1-score evaluated across different training and testing stages. The purple dashed line represents the highest performance achieved at each stage, pointing to the overall declining performance trend across the models (clockwise from top left) BERT, BigBird , Stella + FFNN (RP + Idea) (c), and Stella + FFNN (Idea Only) (d).}
    \label{fig:f1score-comparison-stages}
\end{figure}



We evaluate the performance of various algorithms in idea-source detection (Table \ref{tab:result-table}). Stage 1 achieves the highest F1-score (>90\% for BigBird, Stella) across many different models. We also find, even a simple logistic regression model attains 77\%, suggesting strong lexical cues. Words like \textit{hybrid}, \textit{transformer}, \textit{adaptive}, \textit{intelligently}, \textit{dynamically}, and \textit{advanced} are highly correlated with LLM-generated text.

To further analyze this, we apply Integrated Gradients (IG) Visualization with RoBERTa (Figure \ref{fig:Integ-grad}) \cite{sundararajan2017axiomatic}. IG attributes model predictions by integrating gradients from a baseline input to the actual input, quantifying feature importance. We find terms like \textit{adaptive}, \textit{framework}, \textit{regularized}, and \textit{stability} align with LLMs, likely due to their prevalence in structured academic writing, whereas domain-specific terms like \textit{federated} and \textit{momentum} are more indicative of human ideas.

We observe, BigBird consistently outperforms all other models, leveraging its superior context-length capability to capture both the research problem (RP) and idea representation effectively. Among fine-tuned models (BERT, RoBERTa, T5, BigBird), RoBERTa slightly outperforms BERT, while high-quality embeddings like Stella and GTE surpass idea-only models such as BERT, RoBERTa, and T5 in most stages, highlighting the advantage of robust embedding spaces.


\subsection{Learning Difficulties with the Progression of Praraphrasing Stages}

As training progresses across stages, a consistent decline in performance is observed, as depicted in Figure \ref{fig:f1score-comparison-stages}. When cross-stage train-test is performed, Stage 1 shows a larger degradation, since it contains only 25\% of the data compared to the later stages. In Stages 2 to 5, models generally achieve their highest performance within the stage they were trained on, indicating a strong stage-specific learning effect. Nevertheless, irrespective of cross-stage or within-stage, overall performance declines with the progression of stages. It also indicates that the earlier stages may still retain the ``LLM signature'', which aids detection but gradually diminishes in later stages.

\begin{figure}[htbp]
    \centering
    \subfigure[]{\includegraphics[width=0.48\linewidth]{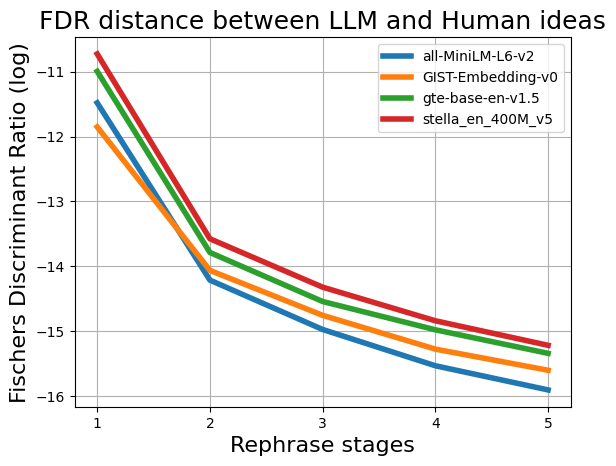}}
    \hfill
 \subfigure[]{\includegraphics[width=0.48\linewidth]{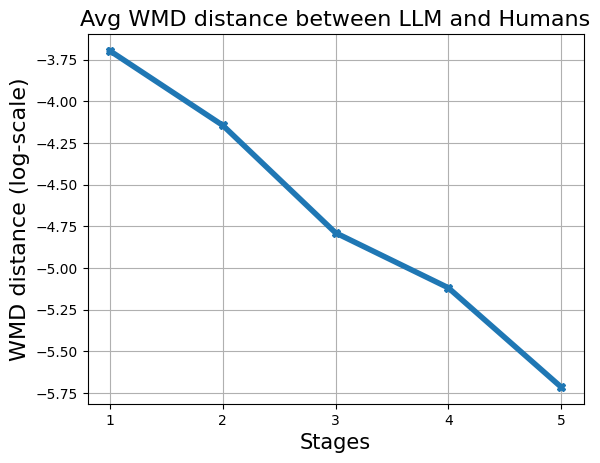}}

    \caption{Visualization of discriminative features between human and LLM-generated ideas. (a) Four different text embedding representations illustrate the decreasing discriminability as we progress from Stage 1 to Stage 5. (b) Word Mover`s Distance also shows a declining trend, indicating reduced differentiation between human and LLM-generated ideas over stages.}
    \label{fig:WMD-FDR}
\end{figure}

To further understand the issue, we investigate the \textbf{Fisher's Discriminant Ratio (FDR)} between LLM and Human ideas acorss different stages \cite{li2014fisher}. 
\[
FDR = \frac{(\mu_1 - \mu_2)^2}{\sigma_1^2 + \sigma_2^2}
\]

where $\mu_1$, $\mu_2$ are the means of the feature (embedding representation) for Human and LLM-generated ideas respectively, and
$\sigma_1^2$, $\sigma_2^2$ are the variances of them respectively. As illustrated in Figure \ref{fig:WMD-FDR}(a), the FDR steadily declines across the stages, irrespective of the embedding representation used. It suggests that as ideas undergo iterative paraphrasing or transformation, their distinguishing characteristics erode, making human and LLM-generated ideas increasingly indistinguishable.

In addition, we examine \textbf{Word Mover’s Distance (WMD)}, a metric that quantifies the effort required to change one document’s word embeddings into another’s, serving as a measure for textual dissimilarity \cite{kusner2015word}. We employ the GloVe-wiki-gigaword-50 embedding model to compute WMD at each stage. Figure \ref{fig:WMD-FDR}(b) reveals a progressive decline in WMD, further reinforcing the notion that the iterative modifications reduce the distinctiveness of LLM-generated content. As the transformation stages accumulate, the `LLM signature' becomes increasingly elusive, making it more challenging to establish a clear boundary between human and LLM-generated ideas.


\begin{figure}
    \centering
    \includegraphics[width=0.4\textwidth]{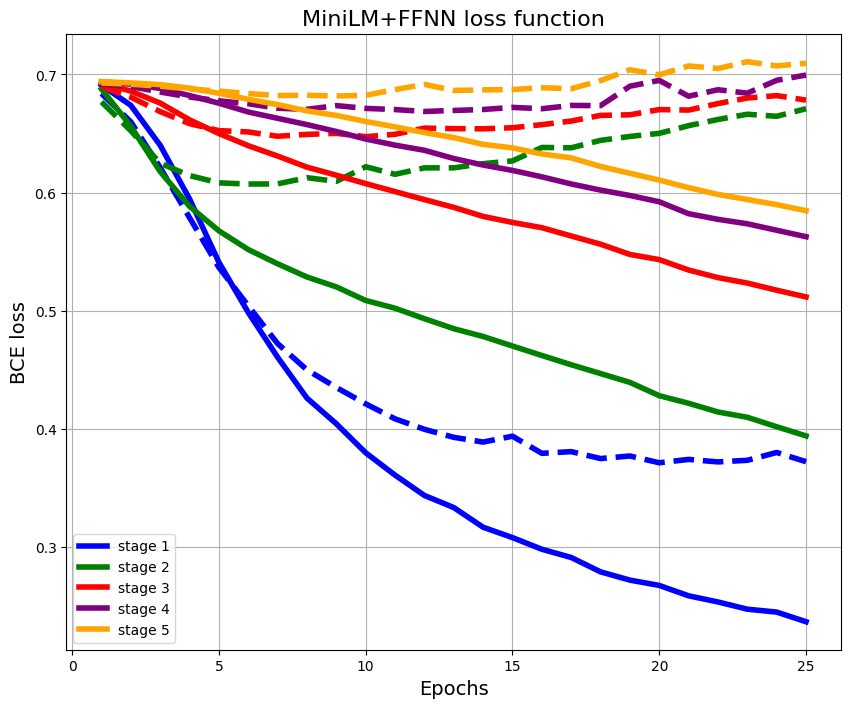}
    \caption{Visualization of the training (solid line) and validation (dashed line) loss curves for the MiniLM + FFNN model across the first 25 epochs, providing insights into learning dynamics and model convergence }
    \label{fig:loss}
\end{figure}

Finally, we investigate the ``learning difficulty'' through the analysis of the loss curves, focusing on the MiniLM + FFNN architecture trained on concatenated (RP + idea) inputs (Figure~\ref{fig:loss}). Computing the average slope of the validation loss across the initial five epochs, given by 
$(L(t+n-1) - L(t))/n$, reveals progressively decreasing slopes of 0.029, 0.014, 0.007, 0.003, and 0.002 from stages 1 to 5. Intuitively, as the cascaded paraphrasing stages progress, this indicates that while early stages rapidly achieve a stable, low-loss plateau, the higher stages quickly plateau at higher losses, followed by an upward drift in validation loss, clearly reflecting increased learning difficulty and poorer generalization .

\subsection{When Does Merging Data Across Stages Help ?}

\begin{figure}[htbp]
    \centering
    \subfigure[]{\includegraphics[width=0.40\textwidth]{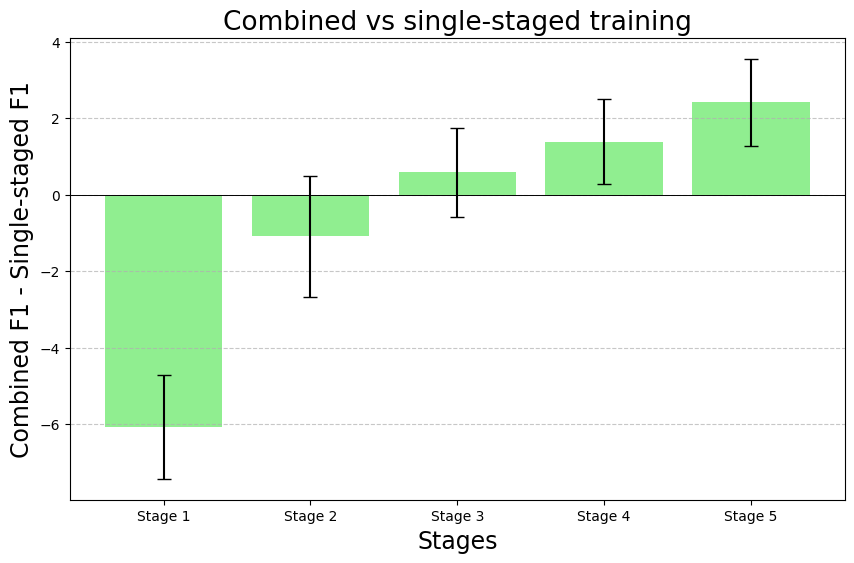}}
    \hfill
 \subfigure[]{\includegraphics[width=0.40\textwidth]{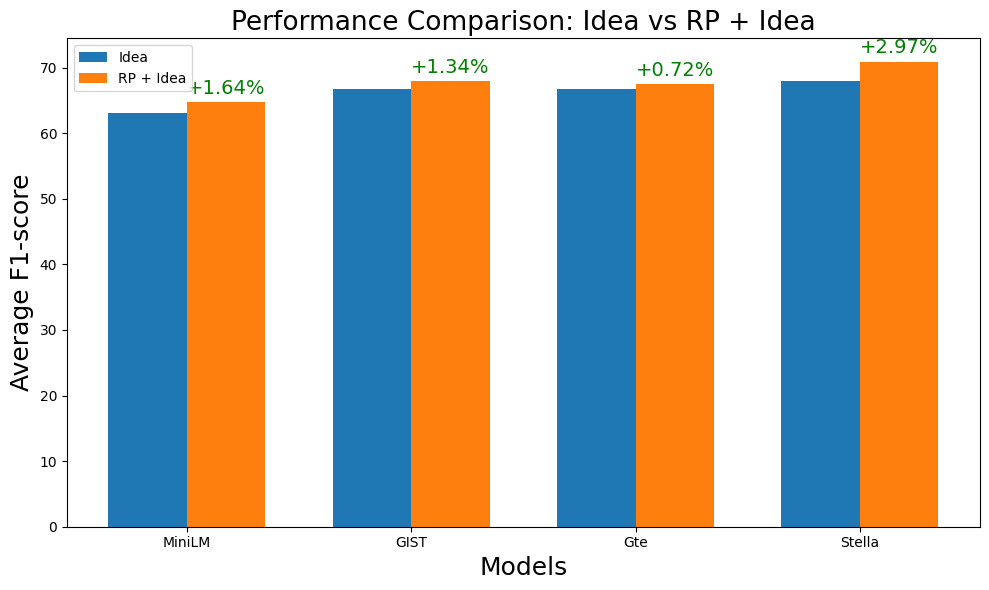}}

    \caption{(a) When combined stages and performed a holistic training, the latter stages get more benfitted, while the earlier stages decline in performance (b) When Research problem embeddings are concatenated with the Idea embedding, we observe a consistent performance increase}
    \label{fig:diff-training}
\end{figure}

We investigate whether combining training examples from different paraphrase stages can improve detection performance, and Figure~\ref{fig:diff-training}(a) reveals that this strategy unexpectedly degrades performance in earlier stages (stages 1 and 2). For stage 1 and 2, combined training declines the average performance by 6.07 and 1.08 points respectively. However, for stage 3, 4, 5, combined training imrpoves the performance by an average of 0.6, 1.4, and 2.4 points respectively, likely due to the increased volume of training data improving generalization. 

In stage 1, even smaller datasets suffice to achieve high accuracy because the LLM’s distinctive ``signature'' remains relatively intact, making it straightforward to distinguish from human-generated content. However, as we progress to later stages (stages 4 and5), repeated paraphrasing gradually erodes these features, creating a more challenging detection task. Under these conditions, adding data from earlier stages proves beneficial because it provides subtle patterns and cues that help the model better learn residual LLM signatures. 


\subsection{Improving Idea Detection by Integrating Problem Context}



Through the embedding + FFNN models, we observe that RP + idea versions of training significantly outperform their idea-only counterparts \ref{fig:diff-training}(b), with observed performance gains of +1.64 (p=0.00) for MiniLM, +1.34 (p=0.04) for GIST, +0.72 (p=0.02) for GTE, and +2.97 (p=0.00) for Stella. 

These gains indicate how incorporating RP helps models learn structured semantic dependencies between RP and their corresponding research idea solutions, thus, leading to richer conceptual representations and reducing ambiguity. In FFNN classifiers, this additional context strengthens decision boundary formation by providing clearer distinctions between different idea categories. However, in the embedding models, RP and ideas are only concatenated at the representation level. To further enhance contextual integration, we plan to explore a cross-attention modeling structure in future work, which may better capture problem-idea interactions and improve the model’s understanding of idea patterns.


\subsection{Simplified Paraphrasing Significantly Reduces Detectability}
\begin{figure}
    \centering
    \includegraphics[width=0.4\textwidth]{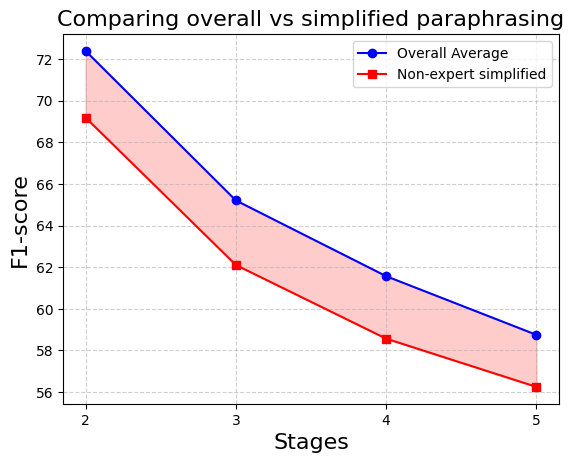}
    \caption{Detection performance (F1-score) consistently deteriorates when ideas are paraphrased into simplified, non-technical language intended for general audiences. }
    \label{fig:nonexpert}
\end{figure}


Across all stages and classifiers, we observe a consistent pattern: simplified paraphrasing intended for non-expert audiences leads to the most substantial reduction in detection performance (Figure \ref{fig:nonexpert}). The average F1-score across all algorithms and paraphrasing stages is 64.5\%, while simplified non-expert paraphrasing underperforms this benchmark significantly by 2.98\% (p-value = 0.03).

This phenomenon likely occurs because non-expert paraphrasing deliberately omits technical nuances, replacing them with simpler, more general language. Such simplification further diminishes the research domain specific linguistic signatures used by models to distinguish between human and LLM-generated ideas. These findings also illustrate critical limitations in current detection algorithms, suggesting they rely heavily on superficial linguistic patterns and struggle to capture deeper `Idea Signature' when technical complexity is removed.

\section{Conclusion}  

This paper examines the ability of SOTA textual ML models to differentiate between human and LLM-generated research ideas, revealing the challenges posed by iterative paraphrasing. Unlike direct text-based detection, idea detection is significantly harder as paraphrasing progressively erodes distinctive LLM signatures, making idea attribution increasingly unreliable. By constructing a systematic dataset from top CS conferences and leveraging advanced LLMs for idea generation and rephrasing, we find that even the best detection models struggle once ideas undergo multiple paraphrasing stages. Our results emphasize that existing classifiers rely heavily on surface-level linguistic features rather than deeply understanding the underlying idea structures, leading to substantial performance declines as paraphrasing progresses.  

In future, we aim to extend this study beyond CS to other scientific disciplines, exploring whether similar challenges persist across diverse knowledge domains. A key direction for improvement involves incorporating the reasoning trajectory of LLMs during idea generation, as tracing the thought process may provide a more robust signal for detection. Additionally, integrating structured knowledge-based embeddings could help models capture deeper conceptual patterns, reducing their dependence on linguistic artifacts and enhancing their ability to differentiate between human and LLM-generated ideas.

\bibliographystyle{acl_natbib}
\bibliography{custom}

@misc{core_portal,
  author    = {{CORE: Computing Research and Education Association of Australasia}},
  title     = {CORE Conference Portal},
  year      = {2025},
  url       = {https://www.core.edu.au/icore-portal},
  note      = {Accessed: 2025-03-09}
}

@article{si2024can,
  title={Can llms generate novel research ideas? a large-scale human study with 100+ nlp researchers},
  author={Si, Chenglei and Yang, Diyi and Hashimoto, Tatsunori},
  journal={arXiv preprint arXiv:2409.04109},
  year={2024}
}

@inproceedings{sundararajan2017axiomatic,
  title={Axiomatic attribution for deep networks},
  author={Sundararajan, Mukund and Taly, Ankur and Yan, Qiqi},
  booktitle={International conference on machine learning},
  pages={3319--3328},
  year={2017},
  organization={PMLR}
}

@article{li2014fisher,
  title={Fisher linear discriminant analysis},
  author={Li, Cheng and Wang, Bingyu},
  journal={CCIS Northeastern University},
  volume={6},
  year={2014}
}

@inproceedings{kusner2015word,
  title={From word embeddings to document distances},
  author={Kusner, Matt and Sun, Yu and Kolkin, Nicholas and Weinberger, Kilian},
  booktitle={International conference on machine learning},
  pages={957--966},
  year={2015},
  organization={PMLR}
}

@article{zhao2023survey,
  title={A survey of large language models},
  author={Zhao, Wayne Xin and Zhou, Kun and Li, Junyi and Tang, Tianyi and Wang, Xiaolei and Hou, Yupeng and Min, Yingqian and Zhang, Beichen and Zhang, Junjie and Dong, Zican and others},
  journal={arXiv preprint arXiv:2303.18223},
  volume={1},
  number={2},
  year={2023}
}

@article{boiko2023emergent,
  title={Emergent autonomous scientific research capabilities of large language models},
  author={Boiko, Daniil A and MacKnight, Robert and Gomes, Gabe},
  journal={arXiv preprint arXiv:2304.05332},
  year={2023}
}

@inproceedings{zhao2023protecting,
  title={Protecting language generation models via invisible watermarking},
  author={Zhao, Xuandong and Wang, Yu-Xiang and Li, Lei},
  booktitle={International Conference on Machine Learning},
  pages={42187--42199},
  year={2023},
  organization={PMLR}
}

@inproceedings{mitchell2023detectgpt,
  title={Detectgpt: Zero-shot machine-generated text detection using probability curvature},
  author={Mitchell, Eric and Lee, Yoonho and Khazatsky, Alexander and Manning, Christopher D and Finn, Chelsea},
  booktitle={International Conference on Machine Learning},
  pages={24950--24962},
  year={2023},
  organization={PMLR}
}

@article{yang2023dna,
  title={Dna-gpt: Divergent n-gram analysis for training-free detection of gpt-generated text},
  author={Yang, Xianjun and Cheng, Wei and Wu, Yue and Petzold, Linda and Wang, William Yang and Chen, Haifeng},
  journal={arXiv preprint arXiv:2305.17359},
  year={2023}
}

@article{hu2023radar,
  title={Radar: Robust ai-text detection via adversarial learning},
  author={Hu, Xiaomeng and Chen, Pin-Yu and Ho, Tsung-Yi},
  journal={Advances in neural information processing systems},
  volume={36},
  pages={15077--15095},
  year={2023}
}

@inproceedings{devlin2019bert,
  title={Bert: Pre-training of deep bidirectional transformers for language understanding},
  author={Devlin, Jacob and Chang, Ming-Wei and Lee, Kenton and Toutanova, Kristina},
  booktitle={Proceedings of the 2019 conference of the North American chapter of the association for computational linguistics: human language technologies, volume 1 (long and short papers)},
  pages={4171--4186},
  year={2019}
}

@article{liu2019roberta,
  title={Roberta: A robustly optimized bert pretraining approach},
  author={Liu, Yinhan and Ott, Myle and Goyal, Naman and Du, Jingfei and Joshi, Mandar and Chen, Danqi and Levy, Omer and Lewis, Mike and Zettlemoyer, Luke and Stoyanov, Veselin},
  journal={arXiv preprint arXiv:1907.11692},
  year={2019}
}

@article{raffel2020exploring,
  title={Exploring the limits of transfer learning with a unified text-to-text transformer},
  author={Raffel, Colin and Shazeer, Noam and Roberts, Adam and Lee, Katherine and Narang, Sharan and Matena, Michael and Zhou, Yanqi and Li, Wei and Liu, Peter J},
  journal={Journal of machine learning research},
  volume={21},
  number={140},
  pages={1--67},
  year={2020}
}

@article{zaheer2020big,
  title={Big bird: Transformers for longer sequences},
  author={Zaheer, Manzil and Guruganesh, Guru and Dubey, Kumar Avinava and Ainslie, Joshua and Alberti, Chris and Ontanon, Santiago and Pham, Philip and Ravula, Anirudh and Wang, Qifan and Yang, Li and others},
  journal={Advances in neural information processing systems},
  volume={33},
  pages={17283--17297},
  year={2020}
}

@article{solatorio2024gistembed,
  title={Gistembed: Guided in-sample selection of training negatives for text embedding fine-tuning},
  author={Solatorio, Aivin V},
  journal={arXiv preprint arXiv:2402.16829},
  year={2024}
}

@misc{zhang2025jasperstelladistillationsota,
      title={Jasper and Stella: distillation of SOTA embedding models}, 
      author={Dun Zhang and Jiacheng Li and Ziyang Zeng and Fulong Wang},
      year={2025},
      eprint={2412.19048},
      archivePrefix={arXiv},
      primaryClass={cs.IR},
      url={https://arxiv.org/abs/2412.19048}, 
}

@article{zhang2024mgte,
  title={mGTE: Generalized Long-Context Text Representation and Reranking Models for Multilingual Text Retrieval},
  author={Zhang, Xin and Zhang, Yanzhao and Long, Dingkun and Xie, Wen and Dai, Ziqi and Tang, Jialong and Lin, Huan and Yang, Baosong and Xie, Pengjun and Huang, Fei and others},
  journal={arXiv preprint arXiv:2407.19669},
  year={2024}
}

@article{wang2020minilm,
  title={Minilm: Deep self-attention distillation for task-agnostic compression of pre-trained transformers},
  author={Wang, Wenhui and Wei, Furu and Dong, Li and Bao, Hangbo and Yang, Nan and Zhou, Ming},
  journal={Advances in neural information processing systems},
  volume={33},
  pages={5776--5788},
  year={2020}
}

@article{baek2024researchagent,
  title={Researchagent: Iterative research idea generation over scientific literature with large language models},
  author={Baek, Jinheon and Jauhar, Sujay Kumar and Cucerzan, Silviu and Hwang, Sung Ju},
  journal={arXiv preprint arXiv:2404.07738},
  year={2024}
}

@article{li2024mlr,
  title={Mlr-copilot: Autonomous machine learning research based on large language models agents},
  author={Li, Ruochen and Patel, Teerth and Wang, Qingyun and Du, Xinya},
  journal={arXiv preprint arXiv:2408.14033},
  year={2024}
}

@article{lu2024ai,
  title={The ai scientist: Towards fully automated open-ended scientific discovery},
  author={Lu, Chris and Lu, Cong and Lange, Robert Tjarko and Foerster, Jakob and Clune, Jeff and Ha, David},
  journal={arXiv preprint arXiv:2408.06292},
  year={2024}
}

@misc{openai2025deepresearch,
  title        = {Introducing Deep Research},
  author       = {OpenAI},
  year         = 2025,
  url          = {https://openai.com/index/introducing-deep-research/},
  note         = {Accessed: 2025-03-14}
}

@misc{OpenAI2025ChatGPT,
  author    = {OpenAI},
  title     = {ChatGPT},
  year      = {2025},
  url       = {https://openai.com/index/chatgpt/},
  note      = {Accessed: March 14, 2025}
}

@misc{Anthropic2025Claude,
  author    = {Anthropic},
  title     = {Claude AI Models},
  year      = {2025},
  url       = {https://docs.anthropic.com/en/docs/about-claude/models/all-models},
  note      = {Accessed: March 14, 2025}
}

@misc{openai2025models,
  author       = {OpenAI},
  title        = {OpenAI Platform - Model Documentation},
  year         = {2025},
  url          = {https://platform.openai.com/docs/models},
  note         = {Accessed: 2025-03-14}
}

@misc{huggingface_mteb2025,
  author       = {Hugging Face},
  title        = {MTEB Leaderboard - Hugging Face Spaces},
  year         = {2025},
  url          = {https://huggingface.co/spaces/mteb/leaderboard},
  note         = {Accessed: 2025-03-14}
}

\clearpage

\appendix

\section{Limitations}  

Our work investigates the diminishing detectability of LLM-generated ideas using SOTA models. Below are the key limitations of our study:  

\begin{itemize}  
    \item \textbf{Limited Domain Research:} We focus on five A* conferences in the Computer Science domain, primarily covering AI and NLP. While these conferences are highly regarded, we do not include other major CS conferences such as CVPR, AAAI, ICSE, or ISSP. Additionally, we exclude papers from other disciplines like Electrical Engineering, Economics, and Psychology. This limits our ability to assess LLMs' idea-generation capacity across a broader range of fields.  

    \item \textbf{Limited Explainability:} While our detection models are state-of-the-art, they lack strong interpretability. Although we incorporate visualizations to analyze their impact, a more explainable approach could provide deeper insights into how these models distinguish human and LLM-generated ideas.  

    \item \textbf{Limited Paraphrasing Stages:} We paraphrase ideas up to stage 5, observing continuous degradation in detectability. However, we do not determine the exact point at which performance drops to chance levels, making LLM-generated ideas entirely indistinguishable.  

    \item \textbf{Black-box LLM:} We generated and paraphrased ideas using SOTA black-box LLMs. However, the field is rapidly evolving, and we could not incorporate some of the latest models, such as GPT-4.5 or Claude-3.7. We also couldnot include LLMs from other platforms like Llama, Mistral, and Gemma. Furthermore, the reliance on black-box models limits interpretability, as we lack insight into their internal mechanisms, decision-making processes, and biases. This restricts our ability to analyze why certain ideas are generated or transformed in specific ways, making it harder to attribute detectability degradation to intrinsic model behaviors versus external linguistic shifts.

\end{itemize}

\section{Paper Statistics}
\label{sec:papers}

We collected a total of 18,581 papers in total as shown in table \ref{tab:accepted_papers_full}
\begin{table}[h]
\centering
\caption{Accepted Papers by Conference and Year}
\begin{tabular}{lccccc}
\cline{1-6}
 & \textbf{2017} & \textbf{2018} & \textbf{2019} & \textbf{2020} & \textbf{2021} \\
 \cline{1-6}
ACL     & 326 & 375 & 310 & 757 & 560 \\
EMNLP   & 345 & 535 & 666 & 726 & 821 \\
ICLR    &  78 & 327 & 489 & 679 & 843 \\
ICML    & 428 & 612 & 759 & 595 & 1164 \\
NeurIPS & 672 & 992 & 1401 & 1852 & 2269 \\
\hline
\end{tabular}
\label{tab:accepted_papers_full}
\end{table}

\section{Prompts Used}
\label{App:Prompts}

\subsection{Prompt for extracting research problem statement}
\label{Appsub:Prompts-problem}

\textit{Act as an experienced researcher with extensive experience in reading and analyzing research papers. You are given the abstract and introduction of a research paper. Your task is to extract the main problem statement the researchers are trying to solve.}

\textit{Guidelines}:

\begin{itemize}
    \item \textit{Focus solely on articulating the main problem, providing only the context necessary to understand the} issue.
    \item \textit{Provide a concise and clear explanation of the problem in exactly one paragraph.}
    \item \textit{Do not summarize, hint at, or include any aspects of the solution provided in the paper.}
    \item \textit{Avoid using phrases such as \"this paper\" or referring to the document itself.}
\end{itemize}

\subsection{Prompt for LLM idea generation}
\label{Appsub:ides-generation}

We used two types of prompting for LLM idea generation. Both approaches take the research problem statement and the instructions as input. 

\textbf{Approach 1 (General)}: \textit{Act as an experienced researcher. Read the problem statement and devise your idea to solve the problem in a detailed manner.Present the research solution in the first person plural (i.e., `we'), as if we are the authors of the idea.}

\textbf{Approach 2 (Detailed)}:\textit{Assume the role of an experienced researcher. You are presented with a problem statement. Your task is to devise a detailed research solution to address the problem. Present the solution as if we are the authors of the idea, using the first-person plural (``we'').
Your response should include the following elements:}

\begin{itemize}
    \item \textit{Clarify the problem clearly to provide proper context.}

    \item \textit{Outline the core idea or approach we propose to solve the problem. Clearly explain the methodology, frameworks, or techniques involved.}

    \item \textit{Provide a step-by-step plan for how we would execute the solution, including any data, tools, or experimental setup required.}

    \item \textit{Highlight the expected results or impact of the solution.}

    \item \textit{Explain why our proposed solution is novel and how it contributes to the field.}

    \item \textit{If there are risks or potential limitation state that briefly}
\end{itemize} 

\textit{Ensure the response is detailed, logical, and well-structured, with researchers in your field as a target audience.}

\subsection{Prompt for Paraphrasing}
\label{Appsub:paraphrase}

\textbf{General Prompting}:

\textit{Given the research problem statement, extract only the main idea from the following \
            research document. \
            Focus solely on the core problem being addressed and the key idea proposed to solve it. \
            Do not include discussions about previous works, results, or any extrapolation. Ensure\
            absolute fidelity to the research document—do not add, remove, or alter any idea.
}

\textbf{For Non-expert Audience}:

\textit{Assume you have a non-expert general audience, who wants to lsiten to your research summary. \
            Now, given the research problem statement,\
            Extract only the main idea from the following research document, and avoid technical jargons. \
            Focus solely on the core problem being addressed and the key idea proposed to solve it. \
            Do not include discussions about previous works, results, or any extrapolation. Ensure\
            absolute fidelity to the research document—do not add, remove, or alter any idea.
            }

\textbf{Brief Summary}:

\textit{Assume you are in a conference and you have an extremely short time to explain your research. \
           Extract only the main idea from the following research document. \
            Focus solely on the core problem being addressed and the key idea proposed to solve it. \
            Do not include discussions about previous works, results, or any extrapolation. Ensure\
            absolute fidelity to the research document—do not add, remove, or alter any idea.}

\textbf{Detailed Technical}

\textit{Assume you are explaining this research to a highly technical audience, such as engineers or scientists.\
            Given the research problem statement, provide a deep, step-by-step breakdown of the research while maintaining\
            a clear and structured explanation. Do not discuss previous work, results, or make any extrapolations. Ensure\
            absolute fidelity to the research document—do not add, remove, or alter any idea.}



\section{Sample LLM response}

\subsection{The paper and the Research problem}

\textbf{The paper:}
Alet, F., Lopez-Contreras, J., Koppel, J., Nye, M., Solar-Lezama, A., Lozano-Perez, T., ... \& Tenenbaum, J. (2021, July). A large-scale benchmark for few-shot program induction and synthesis. \textit{In International Conference on Machine Learning} (pp. 175-186). PMLR.

\subsection{Research Problem}

\begin{tcolorbox}[colback=green!5, colframe=green!50!black, sharp corners]
The primary problem being addressed is the lack of a substantial, real-world dataset for few-shot program induction and synthesis, which is crucial for training machine learning models to generalize effectively from limited examples. While existing methods often rely on synthetic datasets generated from Domain Specific Languages (DSLs), they fail to capture the complexities and nuances of meaningful programs in general-purpose programming languages like C++. The challenge lies in generating a large collection of program induction tasks that satisfy the necessary properties for effective learning, including well-defined input-output examples that are representative of the real-world contexts in which these programs operate.
\end{tcolorbox}

\subsection{Paper and Idea Summary: Stage 1}

\subsubsection{Paper Idea:}
In our research, we address a significant challenge in the field of AI: the need for few-shot program induction and synthesis. Traditional machine learning methods often require vast amounts of data to perform effectively, which is not feasible for the task of program synthesis. The lack of a large-scale benchmark for analyzing few-shot program induction has created a gap in exploring how to train algorithms to generalize from limited examples effectively. By leveraging existing human-generated code and associated unit tests, we aim to dramatically expand the dataset for program induction tasks, thus facilitating more robust learning and generalization for AI systems.

To tackle this issue, we introduce a novel benchmarking dataset named PROG RES. This dataset synthesizes a large collection of program induction tasks by extracting subprograms from real-world codebases (specifically those from competitive programming platforms) and associating them with meaningful input-output examples derived from their unit tests. Our approach focuses on using real programs instead of synthetic data, which significantly enhances the relevance and quality of the training examples. By deriving input-output pairs from subprograms, we provide a structured learning curriculum that reflects the complexities present in practical programming scenarios while keeping the tasks manageable for current machine learning models.

The novelty of our idea lies in its dual focus: not only do we construct a rich benchmark from actual code, but we also innovate by isolating subprograms that can streamline the learning process. This allows for the development of more efficient algorithms capable of tackling complex programming problems with fewer examples. By ensuring that our dataset includes comprehensive tests that cover potential edge cases, we maximize the utility of each task, preparing algorithms to perform robustly in varied real-world programming contexts. Our approach not only enriches the landscape of few-shot learning in program synthesis but also lays a foundational framework for future research in the field, aiming to bridge the gap between human-like programming skills and machine learning capabilities.

\subsubsection{LLM Idea:}

We are addressing a significant shortcoming in the field of few-shot program induction and synthesis—namely, the lack of diverse and rich real-world datasets. Existing synthetic datasets often fall short when it comes to capturing the complexities and variabilities of programming tasks as they exist in actual software projects. Our goal is to bridge this gap by collecting, curating, and augmenting a comprehensive dataset of programming tasks from various real-world contexts. This initiative is particularly focused on programs written in C++, a widely utilized general-purpose programming language that presents unique challenges due to its diverse applications and intricate syntax.

To achieve this, we have developed a systematic approach that includes the rigorous exploration of open-source code repositories such as GitHub and Bitbucket. Using APIs and advanced web scraping techniques, we aim to compile a substantial and varied collection of existing C++ projects that reflect real-world applications ranging from game development to embedded systems. This foundational data collection step is crucial, as it allows us to extract meaningful programming tasks, complete with detailed input-output specifications, constraints, and contextual information—something most current datasets lack. By sourcing these tasks directly from practical coding scenarios, we ensure their relevance and potential for real-world applicability.

What sets our project apart is not only the breadth of our data collection but also the meticulous methodology we employ for task definition, annotation, and validation. We plan to engage domain experts to review our extracted tasks, alongside crowdsourcing efforts to enhance diversity in the dataset. Each task will be carefully structured to include essential details such as example input-output pairs and metadata that highlights the challenges posed in real-world coding scenarios. Furthermore, we aim to enhance our model's few-shot learning capabilities through controlled data augmentation strategies—introducing variability and complexity that better mirrors the real-world coding environment. This unique combination of rigorous data collection, expert validation, and innovative augmentation methodologies positions our approach as a novel and impactful contribution to the landscape of machine learning for program synthesis.

\subsection{Stage 2 rephrases}

\subsubsection{Paper rephrases:}

\textbf{General Rephrase}

The research addresses the lack of a substantial, real-world dataset for few-shot program induction and synthesis by introducing PROG RES, a novel benchmarking dataset that extracts subprograms from real-world codebases with associated unit tests, providing meaningful input-output examples to facilitate more effective machine learning training with limited examples.

\textbf{Simplified Rephrase}

The research focuses on improving how artificial intelligence learns to write computer programs by creating a new, large collection of real-world programming tasks. Instead of using simple, artificially created examples, the researchers gathered actual code and its corresponding tests from competitive programming platforms. This approach provides meaningful and complex examples, allowing AI models to better understand and generate programs even when they have only a few examples to learn from.

\textbf{Brief Rephrase}

The core problem being addressed is the lack of a substantial, real-world dataset for few-shot program induction and synthesis, which is essential for training machine learning models to generalize from limited examples. The key idea proposed to solve this problem is the introduction of a novel benchmarking dataset named PROG RES, which synthesizes program induction tasks by extracting subprograms from real-world codebases and associating them with meaningful input-output examples from unit tests, thus enhancing the relevance and quality of the training data.

\textbf{Detailed Technical Rephrase}

Let me break down this research step-by-step for a technical audience:

1. Core Research Focus
   - Primary objective: Development of a comprehensive dataset for few-shot program induction
   - Target domain: Program synthesis in general-purpose programming languages (C++)
   - Key distinction: Real-world code vs. traditional synthetic datasets

2. Dataset Construction: PROG RES
   - Source material: Competitive programming platforms' codebases
   - Extraction methodology:
     a. Identification and isolation of subprograms from existing codebases
     b. Association with corresponding unit tests
     c. Derivation of input-output pairs from these tests

3. Technical Architecture
   - Data Processing Pipeline:
     a. Codebase parsing and subprogram extraction
     b. Unit test analysis and conversion
     c. Input-output pair generation
   - Quality Assurance:
     - Edge case coverage through comprehensive test sets
     - Validation of subprogram independence
     - Verification of input-output pair completeness

4. Methodological Innovation
   - Subprogram Isolation:
     a. Identification of self-contained code segments
     b. Preservation of functional completeness
     c. Maintenance of real-world complexity while ensuring manageable scope

5. Dataset Properties
   - Composition: Real-world programming constructs
   - Scope: General-purpose programming patterns
   - Granularity: Subprogram level
   - Coverage: Multiple programming paradigms and patterns

6. Technical Requirements
   - Well-defined input-output mappings
   - Comprehensive test coverage
   - Maintainable complexity levels
   - Representative real-world scenarios

7. Implementation Framework
   - Focus on general-purpose programming language (C++)
   - Integration with existing unit test frameworks
   - Structured extraction and validation processes
   - Systematic organization of program induction tasks

This research establishes a structured approach to creating a robust dataset for few-shot program induction, emphasizing real-world applicability while maintaining technical rigor in the extraction and validation processes.

Click to add a cell.

\subsubsection{LLM-idea rephrases:}

\textbf{General Rephrase}

The research aims to address the lack of diverse, real-world datasets for few-shot program induction and synthesis by systematically collecting and curating a comprehensive dataset of C++ programming tasks from open-source repositories. The approach involves using APIs and web scraping to extract meaningful programming tasks, followed by expert validation and data augmentation to create a rich, contextually representative dataset that captures the complexities of real-world coding scenarios.

\textbf{Simplified Rephrase}

The main idea of this research is to create a large and diverse collection of real-world C++ programming tasks. This collection will help train machine learning models to better understand and generate programs by using examples that reflect actual coding scenarios, rather than simplified or artificial ones.

\textbf{Brief Rephrase}

The core problem being addressed is the lack of a substantial, real-world dataset for few-shot program induction and synthesis, which is essential for training effective machine learning models. The key idea proposed to solve this problem is to collect, curate, and augment a comprehensive dataset of programming tasks from various real-world contexts, specifically from open-source C++ projects, ensuring that the tasks reflect the complexities and nuances of real programming scenarios.

\textbf{Detailed Technical Rephrase}

1. Core Research Focus
   - Primary objective: Creation of a comprehensive real-world dataset for few-shot program induction and synthesis
   - Target language: C++
   - Current limitation addressed: Inadequacy of synthetic datasets generated from DSLs

2. Data Collection Architecture
   a) Source identification
      - Primary sources: Open-source repositories (GitHub, Bitbucket)
      - Collection methods: API integration and web scraping techniques
      - Scope: Multiple application domains (game development, embedded systems, etc.)

   b) Data extraction framework
      - Extraction of complete programming tasks
      - Capture of input-output specifications
      - Documentation of operational constraints
      - Preservation of contextual information

3. Methodological Components
   a) Task definition protocol
      - Structured documentation of programming challenges
      - Specification of input-output relationships
      - Constraint documentation
      - Context preservation

   b) Validation framework
      - Expert review system
      - Crowdsourcing integration for diversity enhancement
      - Quality assurance protocols

4. Data Enhancement Strategy
   - Implementation of controlled augmentation techniques
   - Introduction of calculated variability
   - Complexity scaling mechanisms
   - Real-world scenario mapping

5. Technical Requirements
   - Input-output pair documentation
   - Metadata structuring
   - Challenge categorization
   - Context preservation systems

6. Dataset Properties
   - Focus on general-purpose programming language (C++)
   - Real-world application representation
   - Complex syntax handling
   - Diverse application domain coverage

This research architecture specifically addresses the fundamental limitation in current program synthesis systems by creating a dataset that captures the full complexity spectrum of real-world programming scenarios, moving beyond the limitations of synthetic DSL-based datasets.

\end{document}